\documentclass[journal]{IEEEtran}
\usepackage{graphicx,float,stfloats,amssymb,multirow,pifont,amsmath,scalerel,makecell}
\usepackage{bm,hyperref,xcolor,enumitem,booktabs,arydshln,footnote,url}
\usepackage[caption=false]{subfig}
\usepackage{color}
\usepackage{amsthm,amsmath,amssymb}
\usepackage{mathrsfs}
\usepackage{threeparttable}

\usepackage[subfigure,titles]{tocloft}
\usepackage[numbers]{natbib}
\hypersetup{hypertex=true,
	colorlinks=true,
	linkcolor=blue,
	anchorcolor=blue,
	citecolor=blue}

\graphicspath{{figs/}}

\begin{document}
\title{Kolmogorov-Arnold Network for Remote Sensing Image Semantic Segmentation}
\author{Xianping~Ma,
				Ziyao~Wang,
				Yin~Hu,
        Xiaokang~Zhang,~\IEEEmembership{Senior Member,~IEEE,}
        Man-On~Pun,~\IEEEmembership{Senior Member,~IEEE}
		\thanks{This work was supported in part by the National Natural Science Foundation of China under Grant 42371374 and 41801323, in part by the Guangdong Provincial Key Laboratory of Future Networks of Intelligence under Grant 2022B1212010001, and in part by the Guangdong Basic and Applied Basic Research Foundation under Grant 2024A1515010454. \textit{(Corresponding authors: Man-On Pun)}}
		\thanks{Xianping Ma, and Man-On Pun are with the School of Science and Engineering, The Chinese University of Hong Kong, Shenzhen, Shenzhen 518172, China (e-mails: xianpingma@link.cuhk.edu.cn; ziyaowang2@link.cuhk.edu.cn; yinhucuhksz@gmail.com; SimonPun@cuhk.edu.cn).}
		\thanks{Xiaokang Zhang is with the School of Information Science and Engineering, Wuhan University of Science and Technology, Wuhan 430081, China.(e-mail: natezhangxk@gmail.com).}
	}
\maketitle
\begin{abstract}
Semantic segmentation plays a crucial role in remote sensing applications, where the accurate extraction and representation of features are essential for high-quality results. Despite the widespread use of encoder-decoder architectures, existing methods often struggle with fully utilizing the high-dimensional features extracted by the encoder and efficiently recovering detailed information during decoding. To address these problems, we propose a novel semantic segmentation network, namely DeepKANSeg, including two key innovations based on the emerging Kolmogorov–Arnold Network (KAN). Notably, the advantage of KAN lies in its ability to decompose high-dimensional complex functions into univariate transformations, enabling efficient and flexible representation of intricate relationships in data. First, we introduce a KAN-based deep feature refinement module, namely DeepKAN to effectively capture complex spatial and rich semantic relationships from high-dimensional features. Second, we replace the traditional multi-layer perceptron (MLP) layers in the global-local combined decoder with KAN-based linear layers, namely GLKAN. This module enhances the decoder’s ability to capture fine-grained details during decoding. To evaluate the effectiveness of the proposed method, experiments are conducted on two well-known fine-resolution remote sensing benchmark datasets, namely ISPRS Vaihingen and ISPRS Potsdam. The results demonstrate that the KAN-enhanced segmentation model achieves superior performance in terms of accuracy compared to state-of-the-art methods. They highlight the potential of KANs as a powerful alternative to traditional architectures in semantic segmentation tasks. Moreover, the explicit univariate decomposition provides improved interpretability, which is particularly beneficial for applications requiring explainable learning in remote sensing. The source code for this work will be accessible at \href{https://github.com/sstary/SSRS}{https://github.com/sstary/SSRS}.
\end{abstract}


\IEEEpeerreviewmaketitle

\section{Introduction}\label{sec:intro}
Semantic segmentation of remote sensing image is a fundamental task in Earth observation, forming the foundation for various downstream applications, such as environmental monitoring \cite{yuan2020deep, cao2022coarse, li2024cross}, disaster management \cite{khan2021deepsmoke, huang2022evaluation, bo2022basnet}, and urban planning \cite{zhang2022extraction, fu2022role, alghamdi2024smart}. In recent years, to achieve efficient and accurate semantic segmentation, researchers have increasingly turned to deep learning methods for the automatic interpretation of remote sensing data. Specifically, early studies predominantly employed Convolutional Neural Networks (CNNs), which proved effective in capturing local features and spatial hierarchies in remote sensing images. Subsequently, the Transformer-based approach has been widely developed due to its unique attention mechanism and strong long-range contextual dependencies capability. More recently, the Mamba model has garnered attention for its low computational complexity in remote sensing image semantic segmentation tasks. The adoption of these deep learning-based methods has significantly improved the performance of semantic segmentation, enabling more robust and accurate analysis of remote sensing images.

Initially, many researchers have proposed methods based on CNNs to address the challenges of high-resolution remote sensing image semantic segmentation. Among these, Long et al. \cite{long2015fully} introduced the Fully Convolutional Network (FCN), which allowed CNN-based architectures to perform dense predictions, achieving true pixel-level semantic segmentation. Subsequently, the UNet model was proposed by Ronneberger et al. \cite{ronneberger2015u}, which features a symmetric encoder-decoder architecture and utilizes skip connections to mitigate feature loss during the forward propagation process, thereby improving segmentation accuracy. More recently, a significant number of CNN-based approaches have been proposed to tackle remote sensing semantic segmentation problems \cite{zheng2023self, naseer2024holistic, adugna2024assessing}. At the same time, with an increasing focus on understanding the working mechanisms of these models, it has become widely accepted in the research community that CNNs' success in dense prediction tasks is largely due to their convolutional operations, which focus on local features \cite{Dusmanu_2019_CVPR, li2021survey, khan2020survey}. However, a major challenge remains in capturing distant contextual dependencies effectively, which limits their ability to model long-range interactions and global context information.

Fortunately, Transformer-based models have been introduced into the field of computer vision \cite{dosovitskiy2020image, liu2021swin, ranftl2021vision}. They utilize the multi-head self-attention to capture global contextual information in images, thereby modeling long-range dependencies. This unique attention mechanism alleviates the challenge faced by CNNs in capturing context information. In the remote sensing filed, several studies have explored the application of Transformer-based models in remote sensing image semantic segmentation tasks up to now \cite{li2021abcnet, wang2022unetformer, wu2023cmtfnet, ma2024frequency}, achieving notable successes. While Transformers are known for their powerful ability to capture global context, their computational process demands substantial memory and computational resources. As a result, researchers are often forced to make trade-offs between the computational cost of the model and its feature extraction capability, which limits the practical application of Transformer models in certain scenarios.

More recently, the Mamba model \cite{gu2024mamba, liu2024vmamba}, which is theoretically based on the state space model (SSM), has attracted much attention because of its ability to model long-distance relationships with linear complexity \cite{gu2024mamba}. In the field of remote sensing, RSMamba \cite{chen2024rsmamba}, RS-Mamba \cite{zhao2024rs} and RS$^{3}$Mamba \cite{ma2024rs} introduce Mamba into remote sensing image semantic segmentation for the first time. Later, PyramidMamba \cite{wang2024pyramidmamba} and PPMamba \cite{hu2024ppmamba} integrate pyramid pooling mechanism and selective scan module, proving that this hybrid structure effectively captures both local and global features of high-resolution remote sensing images. As a novel model architecture, Mamba still holds significant potential that remains to be further explored. Unfortunately, similar to CNN-based and Transformer-based approaches, most existing deep learning methods rely on MLP-based calculations, which are limited in handling complex high-dimensional features. More specifically, MLP relies on global linear transformations and activation functions for feature processing, which can struggle to effectively capture intricate local and non-linear relationships in high-dimensional features. However, in complex high-resolution remote sensing image tasks, the processing of high-dimensional features and the restoration of semantic features rely heavily on fine-grained operations.

To address the aforementioned challenges, we introduce trending KAN technology, which leverages the Kolmogorov–Arnold representation theorem to decompose high-dimensional functions into a combination of univariate functions, offering superior capability in modeling complex and localized feature relationships \cite{liu2024kan, liu2024kan2}. Meanwhile, considering the importance of pre-trained models for feature extraction, we introduce KAN from two perspectives: high-dimensional feature refinement and semantic feature decoding. Specifically, we propose a DeepKAN module to further extract the high-dimensional semantic information generated by the encoder. Through non-linear operations along the channel dimension, the rich semantic information embedded in the abstract high-dimensional features can be effectively exploited. In addition, we present a GLKAN module that replaces the MLP operations in the global-local enhancement decoder with KAN-based operations. This facilitates the stepwise restoration of semantic information, ultimately achieving precise pixel-level class predictions. The fourfold contributions are summarized in the following:
\begin{itemize}
\item A KAN-based deep feature refinement module, namely DeepKAN, is proposed to further explore the rich high-dimensional semantic space of remote sensing data. 
\item A GLKAN-based decoder is introduced to progressively restore ground category information through more concise non-linear control. 
\item Capitalizing on the DeepKAN and the GLKAN-based decoder, a novel semantic segmentation network, namely DeepKANSeg is proposed. 
\item Comprehensive experiments on two widely used fine-resolution remote sensing datasets and two representative encoders confirmed the superior performance of the KAN-based method.
\end{itemize}
The adoption of KAN in this work is expected to pave the way for a novel and highly promising research direction, offering substantial potential for further advancements in this field. The structure of the paper is as follows: Sec.~\ref{sec:rel} reviews related work on semantic segmentation and the KAN in remote sensing. Sec.~\ref{sec:met} introduces the proposed DeepKANSeg, and provides a detailed explanation of its two key components. Sec.~\ref{sec:exp} presents a comprehensive analysis of the extensive experiments along with a prospective discussion on future developments. Finally, Sec.~\ref{sec:con} concludes this work.

\section{Related Works}\label{sec:rel}
\subsection{Remote Sensing Semantic Segmentation}
Semantic segmentation has achieved remarkable advancements in the field of remote sensing. Convolutional neural networks (CNNs) have served as foundational tools in remote sensing image analysis, owing to their robust feature extraction capabilities \cite{kotaridis2021remote, yuan2021review}. Models such as PPResNet \cite{bo2018cnn}, DeepResUnet \cite{yi2019semantic}, and ResUNet-a \cite{diak2020resunet} have demonstrated strong performance in semantic segmentation tasks for remote sensing images. These approaches systematically examined the effects of various convolutional kernels and network depths on task performance. In particular, ResUNet-a \cite{diak2020resunet} introduced a conditioned relationship between different tasks, enhancing the model's convergence efficiency. Despite their success, CNNs are inherently limited by their local receptive field and inability to effectively capture global contextual information. This limitation can result in significant information loss when processing complex remote sensing scenarios, particularly in large-scale or high-resolution images.

To overcome the challenge of CNNs lacking long-range dependency, the Transformer architecture has been introduced in recent years to effectively capture global information, making it particularly well-suited for processing large-scale images in remote sensing semantic segmentation \cite{alei2023transformers, xu2021efficient, he2022swin, ma2022crossmodal, wang2022unetformer, wu2023cmtfnet, ma2024multilevel, yao2024ssnet}. For example, UNetFormer \cite{wang2022unetformer} incorporated an efficient global-local attention mechanism to enhance the decoder’s ability to interpret global and local information. Similarly, FTransUNet \cite{ma2024multilevel} addressed the multimodal fusion challenge by combining a fusion vision Transformer with CNN blocks to extract local features across different modalities. Despite these advancements, the high computational complexity of Transformer models limits their widespread application in high-resolution remote sensing imagery, especially in resource-constrained environments.

Mamba is an emerging lightweight deep learning framework that leverages the state space model to balance computational efficiency and accuracy \cite{liu2024vision, zhu2024vision, xu2024survey}, making it well-suited for remote sensing semantic segmentation tasks in resource-constrained environments \cite{chen2024rsmamba, zhao2024rs, ma2024rs, zhu2024samba, chen2024changemamba}. For instance, RS-Mamba \cite{zhao2024rs} introduced an omnidirectional selective scan technique to capture large spatial features from multiple directions, expanding the scanning capabilities of Mamba for a more comprehensive and effective learning process. Meanwhile, RS$^{3}$Mamba \cite{ma2024rs} used auxiliary CNN layers to extract local features while using Mamba to model long-range dependencies. These approaches highlight two distinct strategies for introducing Mamba into remote sensing applications.

In general, the aforementioned structures often utilize multi-layer perceptron (MLP) as a fundamental component. Recently, the Kolmogorov–Arnold Network (KAN) has been proposed as a promising alternative to MLP, offering the potential to introduce new learning capabilities \cite{liu2024kan, liu2024kan2, somvanshi2024survey}. However, there is currently no research exploring the application of KAN to remote sensing semantic segmentation for high-resolution imagery, which presents a significant direction for future investigation. As a potential alternative, the efficacy of KAN in enhancing semantic segmentation for remote sensing data remains to be validated through further experiments and studies.

\subsection{Kolmogorov–Arnold Network}
KAN \cite{liu2024kan} is a novel neural network architecture designed to replace the traditional MLP structure by utilizing a learnable function in place of a fixed activation function, leveraging the Kolmogorov–Arnold theorem. This theorem states that any multivariable continuous function can be represented as a combination of a finite number of single-variable continuous functions. KAN operationalizes this by decomposing complex high-dimensional functions into a set of simple one-dimensional functions, typically parameterized as spline functions. This approach offers extremely high flexibility and enables the modeling of intricate functions, while is also anticipated to enhance the model's interpretability.

KAN has been initially applied in the fields of computer vision \cite{cheon2024demonstrating} and medical imaging \cite{li2024u, zeng2024multilevel}, demonstrating its versatility and effectiveness. In computer vision, Convolutional KAN \cite{bodner2024convolutional} integrated KAN into convolutional layers by implementing spline functions as kernels. This approach outperforms traditional convolutional layers, resulting in a more adaptable and learnable network. In medical imaging, U-KAN \cite{li2024u} incorporated KAN layers into the U-Net framework. By introducing non-linear learnable activation functions, it achieved improved accuracy and interpretability. In addition, TransUKAN \cite{wu2024transukan} leveraged KAN to capture non-linear relationships with minimal additional parameters, addressing the Transformer’s limitations in extracting local information while reducing memory usage and computational overhead.

In the field of remote sensing, the application of KAN is still in its early stages. Preliminary studies highlight the potential of KAN in remote sensing optical image processing \cite{cheon2024kolmogorov, liu2024aekan} and hyperspectral image classification \cite{jamali2024learn, lobanov2024hyperkan, seydi2024unveiling, wang2024spectralkan}. In particular, Cheon \cite{cheon2024kolmogorov} replaced traditional MLP layers with KAN for remote sensing image classification and explored its combination with various pre-trained CNN and ViT models. The results reveal that KAN enhances classification accuracy while reducing computational complexity. Furthermore, AEKAN \cite{liu2024aekan} marked the first use of pure KAN in remote sensing multimodal change detection tasks, with experimental results demonstrating its effectiveness and superiority. Wav-KAN \cite{seydi2024unveiling} introduced wavelet functions as learnable activation mechanisms, facilitating the non-linear mapping of input spectral signatures. Although existing studies have demonstrated the potential of KAN in remote sensing, its application in this field remains in the exploratory stage. Most current research focuses on straightforward image classification tasks, leaving the full potential of KAN-based models for more complex remote sensing applications largely untapped. Consequently, exploring effective ways to integrate KAN with deep semantic refinement and restoration in segmentation tasks remains an important and promising area for further investigation.

\begin{figure}[t]
	\centering
	{\includegraphics[width=0.95\linewidth]{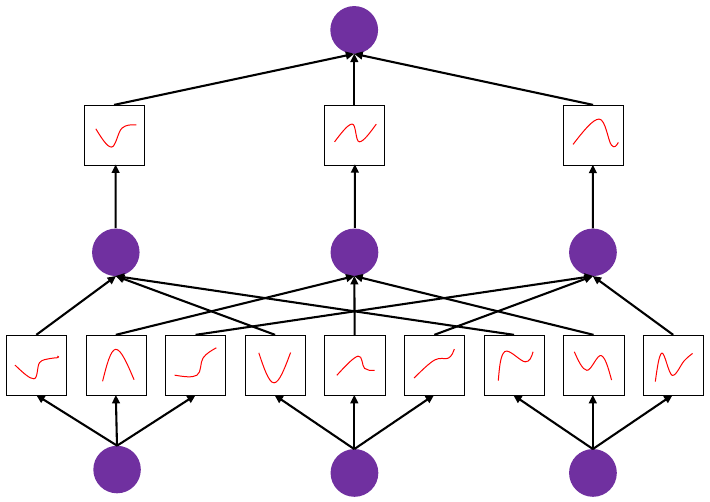}}
	\caption{The structure of a two-layer KAN follows Eq. \ref{eq1}. It learns through multiple learnable activation functions.}
	\label{fig_KANLayer}
\end{figure}

\section{Methodology}\label{sec:met}
\subsection{Preliminary: KAN}
This paper aims to integrate the Kolmogorov–Arnold Network (KAN) into the general framework for remote sensing semantic segmentation, building upon the demonstrated efficiency and interpretability of KAN as highlighted in \cite{liu2024kan}. KAN is inspired by the Kolmogorov–Arnold representation theorem, which provides a theoretical foundation for expressing any multivariable continuous function as a combination of single-variable continuous functions \cite{braun2009constructive}. This theorem is expressed as:
\begin{equation}
    f(\textbf{x}) = \sum_{q} \phi_q\left(\sum_{p} \psi_{pq}(x_p)\right),\label{eq1}
\end{equation}
where $\textbf{x}$ represents the input vector, $ \phi_q $ and $\psi_{pq}$ are continuous single-variable functions. The process is also illustrated in Fig.\ref{fig_KANLayer}. In contrast to MLPs, which rely on fully connected layers with fixed activation functions, KAN stands out by utilizing learnable activation functions and parameterized activation functions as weights, thereby removing the need for traditional linear weight matrices. By parameterizing $ \phi_q $ and $\psi_{pq}$ as spline functions, KAN facilitates highly flexible modeling of intricate relationships. In addition, the explicit univariate decomposition provides improved interpretability, which is expected to offer deeper insights into the model's internal decision-making processes.

Deep KAN extends the foundational principles of KAN to deeper architectures, enabling the modeling of hierarchical and intricate relationships in high-dimensional data. The cornerstone of deep KAN lies in its utilization of learnable activation functions, hereby a matrix of one-dimensional functions can be used to represent the network layer of KAN (KAN layer), which can be defined as:
\begin{equation}
    \bm{\Phi} =\{\phi_{q,p}\}, \qquad p = 1,2,\cdots,n_{\rm{in}},~q=1,2,\cdots,n_{\rm{out}},\label{eq2}
\end{equation}
where $n_{\rm{in}}$ denotes the dimension of input features, and $n_{\rm{out}}$ represents the dimension of output features. Then the transformation of the deep KAN network from layer $k$ to layer $k+1$ can be expressed in matrix form as: $\textbf{x}_{k+1} = \bm{\Phi}_{k}\textbf{x}_{k}$. Finally, a $\emph{K}$-layer deep KAN can be characterized as a composition of multiple KAN layers, represented as:
\begin{equation}
    \rm{KAN}(\textbf{x}) = (\bm{\Phi}_{\emph{K}-1}\circ\bm{\Phi}_{\emph{K}-2}\circ\cdots\circ\bm{\Phi}_{1}\circ\bm{\Phi}_{0})\textbf{x}.\label{eq3}
\end{equation}
This module can be seamlessly integrated into existing semantic segmentation networks for remote sensing tasks, offering a powerful solution to enhance feature representation learning.

\begin{figure*}[t]
	\centering
	{\includegraphics[width=0.95\linewidth]{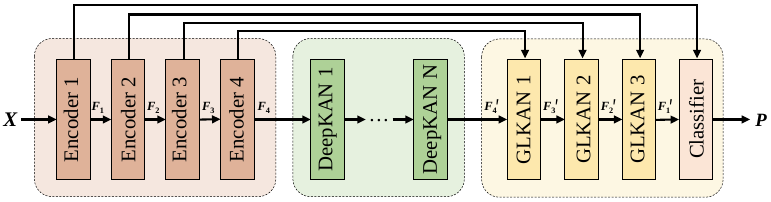}}
	\caption{The overview of the proposed DeepKANSeg, which is comprised of three parts: an encoder, a deep feature refinement module, and a decoder. The combination of a pre-trained encoder with the KAN-based deep feature refinement module and decoder further enhances the semantic segmentation performance.}
	\label{fig_framework}
\end{figure*}

\subsection{Proposed DeepKANSeg}
In this paper, we introduce KAN into general semantic segmentation methods for remote sensing by developing an innovative framework based on the classical encoder-decoder architecture. The proposed DeepKANSeg, illustrated in Fig.~\ref{fig_framework}, consists of three primary components: an encoder, a deep feature refinement module, and a decoder. Given the absence of pre-trained models based on KAN, which could significantly hinder segmentation performance, the well-established CNN or Transformer-based encoders are utilized. During both the training and test stages, the encoder extracts multi-scale features from the input. These features are subsequently processed by the deep feature refinement module, which leverages stacked DeepKAN modules to effectively exploit high-dimensional abstract representations. The refined features are then passed through a decoder comprising GLKAN modules, which iteratively reconstruct semantic information with corresponding skip connections from the encoder. Notably, the weighted sum for handling skip connections is omitted for the sake of clarity in the figures. Finally, a classifier generates the pixel-level segmentation map. 

\begin{figure*}[t]
	\centering
	{\includegraphics[width=0.95\linewidth]{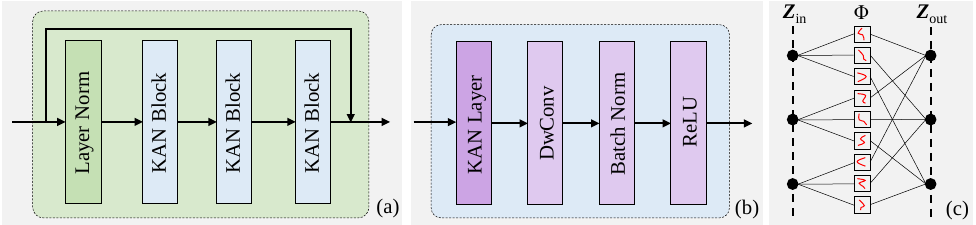}}
	\caption{(a) The structure of the proposed DeepKAN, (b) the structure of the core refinement module KAN block, (c) and a simplified illustration of the KAN layer. The high-dimension feature is refined by the stacked KAN layer, which is crucial for extracting complex remote sensing semantic information.}
	\label{fig_DeepKAN}
\end{figure*}

\subsection{Encoder}
Given the input denoted as $\bm{X} \in \mathbb{R}^{H \times W \times C}$, where $H, W$ are the spatial dimensions, and $C$ is the number of channels (e.g., $C=3$ for RGB images), the encoder progressively extracts feature maps with different scales, denoted as $\bm{F}_i \in \mathbb{R}^{\frac{H}{2^{i+1}} \times \frac{W}{2^{i+1}} \times C_i}$, where $C_i$ is the number of channels at stage $i = \{1, 2, 3, 4\}$. In this work, CNN-based ResNet \cite{He_2016_CVPR} and Transformer-based plain ViT \cite{vit, li2022exploring} are utilized to demonstrate the flexibility and robustness of the proposed DeepKANSeg.

\textbf{ResNet:} The input is initially processed through a convolutional layer, followed by a max-pooling operation that reduces the spatial dimensions. The core network consists of four stages of residual blocks, where each stage doubles the channel dimensions and reduces the spatial resolution using convolutions. Each residual block contains several convolutional layers with skip connections, and they finally produce multi-scale feature maps \( \bm{F}_i \).

\textbf{Plain ViT:} The input is divided into non-overlapping patches of size $P \times P$. Each patch is flattened into a vector, forming a sequence of patch embeddings $\bm{Z}_0 \in \mathbb{R}^{L \times D}$, where $L = \frac{H \times W}{P^2}$ represents the number of patches, and $D$ is the embedding dimension. Positional encodings are added to retain spatial information. The core network consists of a series of transformer layers, each comprising a multi-head self-attention (MSA) module and a feed-forward network (FFN), both equipped with layer normalization and residual connections. The output of the final transformer layer is reshaped to $\bm{F}_p \in \mathbb{R}^{\frac{H}{P} \times \frac{W}{P} \times D}$ and then used to generate multi-scale feature maps through pyramid modules, which includes parallel convolutions or deconvolutions \cite{li2022exploring}. This simple approach effectively generates multi-scale features $\bm{F}_i$.

\subsection{DeepKAN}
The high-dimensional feature $\bm{F}_{\rm{4}}$ extracted by the encoder, is processed through the proposed deep feature refinement module composed of stacked DeepKAN modules. It can effectively learn abstract semantic features through fine-grained learnable activation functions. Initially, $\bm{F}_{\rm{4}}$ is flattened into serialized data $\bm{Z}_{\rm{in}}$, and then passed through $\rm{N}$ DeepKAN modules. Each DeepKAN module consists of a layer normalization and three KAN blocks, as shown in Fig.~\ref{fig_DeepKAN}(a), where the KAN block forms the primary computational units. Considering the richness of high-dimensional channel information, a depth-wise convolution operation along the channel dimension is employed to assist the learning process of KAN. More specifically, a KAN block includes a KAN layer, an efficient depth-wise convolutional layer (DwConv), a batch normalization layer (BN), and a ReLU activation function, as depicted in Fig.~\ref{fig_DeepKAN}(b). The process of KAN block, denoted as $\mathcal{F}(\cdot)$, can be expressed as:
\begin{equation}
    \mathcal{F}(\bm{Z}_{\rm{in}})=\sigma({\rm BN}({\rm DwConv}(\Phi\bm{Z}_{\rm{in}}))),\label{eq4}
\end{equation}
where $\sigma$ represents the ReLU activation, and $\Phi$ denotes the KAN layer as defined in Eq. \ref{eq2}. Fig.~\ref{fig_DeepKAN}(c) provides a simplified illustration of $\Phi$. The output $\bm{Z}_{\rm{out}}$ serves as the input for the subsequent KAN block. After processing through $N$ DeepKAN modules, the final $\bm{Z}_{\rm{out}}$ is reshaped back to the original input size denoted as $\bm{F}^{\prime}_{4}$. At this stage, the high-dimensional semantic information is further refined and enriched.

\subsection{Decoder}
\begin{figure}[t]
	\centering
	{\includegraphics[width=\linewidth]{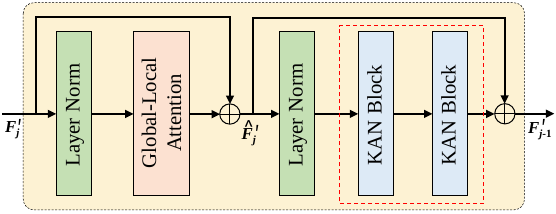}}
	\caption{The structure of the proposed GLKAN. It resembles the classic transformer block, with the component highlighted in the red box modified into KAN-based blocks.}
	\label{fig_GLKAN}
\end{figure}

The output from the deep feature refinement module is subsequently decoded using the proposed decoder, with GLKAN as its core component. As illustrated in Fig.~\ref{fig_GLKAN}, GLKAN adopts the structure of a typical transformer block, comprising an attention block and a fully connected block, each equipped with a layer normalization (LN). The attention block employs a classic Global-Local Attention mechanism \cite{wang2022unetformer}, which consists of two parallel branches: a global branch leveraging window-based self-attention to capture long-range dependencies \cite{he2022swin} and a local branch using convolutional operations for local feature exploiting. The outputs of these branches are fused via element-wise summation, enabling comprehensive extraction of both global and local contexts. GLKAN differs from the standard Global-Local transformer block by substituting the fully connected layer with KAN-based modules, allowing fine-grained channel-level operations to more effectively recover semantic features. The process of GLKAN can be elaborated as:
\begin{equation}
\begin{aligned}
    &\bm{\hat{F}}^{\prime}_{j}={\rm GLAttn}({\rm LN}(\bm{F}^{\prime}_{j}))+\bm{F}^{\prime}_{j},\\
    &\bm{F}^{\prime}_{j\rm{-1}}=\mathcal{F}(\mathcal{F}({\rm LN}(\bm{\hat{F}}^{\prime}_{j})))+\bm{\hat{F}}^{\prime}_{j},\label{eq5}
\end{aligned}
\end{equation}
where ${\rm GLAttn}$ represents the Global-Local Attention mechanism, $\mathcal{F}(\cdot)$ denotes the KAN block as defiend in Eq. \ref{eq4}, and the subscript $j = \{4, 3, 2\}$ is the number of channels at decoder stage. The decoder progressively decodes the features, gradually recovering spatial information and revealing ground class details. Finally, $\bm{F}^{\prime}_{1}$ is processed by a classifier to generate the segmentation map denoted by $\bm{P} \in \mathbb{R}^{H \times W \times C^{\prime}}$, where $C^{\prime}$ is the number of classes.

\begin{table}[h]\scriptsize
	\centering
	\caption{Illustration of two datasets.}
	\setlength{\tabcolsep}{5mm}{
		\begin{tabular}{m{2.0cm}<{\centering}|m{2.0cm}<{\centering}|m{2.0cm}<{\centering}}
			\hline
			&  \textbf{ISPRS Vaihingen} & \textbf{ISPRS Potsdam}  \\
			\hline
			Geographical Type & Small Town & Urban\\
			\hline
			Spatial Resolution & 9 cm & 5 cm\\
			\hline
			Spectral Channel & Near-InfraRed, Red and Green (NIRRG) & Red, Green and Blue (RGB) \\
			\hline
			Data Volume & 16 patches & 24 patches \\
			\hline
			Patch Size &	Around $2500 \times 2000$ & $6000 \times 6000$ \\
			\hline
			Training/Test Sample Size & 256 & 256 \\
			\hline
			Test Stride & 32 & 128 \\
			\hline
			Training-Test Ratio & 3-1 & 3-1 \\
			\hline
			Training-Test Sample Volume & 960-320 & 10368-3456 \\
			\hline
			Proportion of Clutter & Only 1\%	& Around 5\% \\
			\hline
	\end{tabular}}\label{tab:datasets}
\end{table}

\begin{figure}[t]
\centering
{\includegraphics[width=0.98\linewidth]{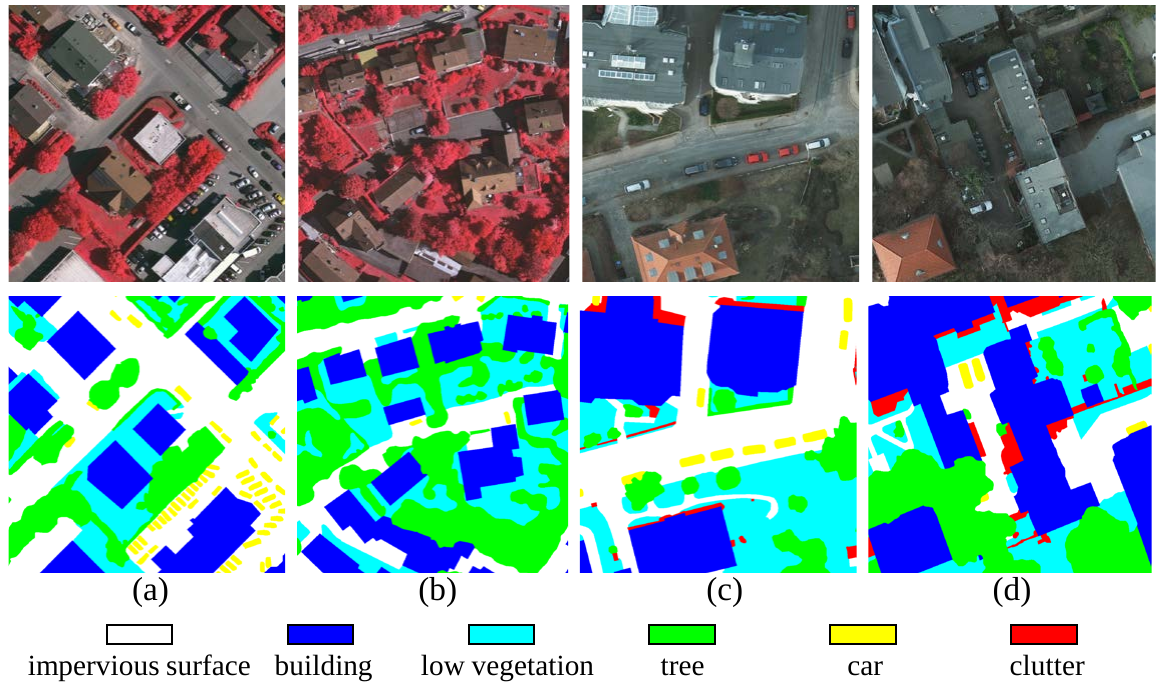}}
\caption{Four visual samples of size $1024 \times 1024$ from the Vaihingen (first two columns) and Potsdam (last two columns) datasets, respectively, are presented. The first row contains orthophotos, with Vaihingen represented in NIRRG channels and Potsdam in RGB channels, while the second row shows the corresponding ground truth. The urban environments in these datasets, with their diverse and complex features, make semantic segmentation a particularly challenging task. Experiments conducted on these two datasets provide a comprehensive validation of the model's performance.}
\label{data_show}
\end{figure}

\section{Experiments and Discussion}\label{sec:exp}
\subsection{Datasets}
The ISPRS Vaihingen and Potsdam datasets both consist of True Orthophotos, which are obtained through aerial photography. These high-resolution datasets cover large areas and depict intricate ground environments, featuring a variety of land cover types including five foreground classes: {\em Impervious Surface (Imp.)}, {\em Building (Bui.)}, {\em Low Vegetation (Low.)}, {\em Tree (Tre.)} and {\em Car}. Their details are summarized in Table~\ref{tab:datasets}. To improve reading efficiency, we adopt a sliding window mechanism to extract small samples from the complete patches. The training and test samples for both datasets are $256$, while considering the difference in data volume, we set the sliding window stride to $32$ and $128$ for the two datasets during the test stage, respectively. 

The two datasets differ significantly in terms of geographical types, spatial resolution, spectral information, and data volume. The complexity and variety of urban and small-town scenes present challenges for developing robust algorithms capable of accurately segmenting and classifying diverse objects. Figure~\ref{data_show} illustrates four data samples from both datasets. Therefore, these datasets are invaluable for evaluating the performance of deep learning models in complex environmental settings, making performance experiments on them both reliable and insightful.

\subsection{Evaluation Metrics}
To evaluate the segmentation performance of the proposed DeepKANSeg, we utilize the mean ${\rm F_1}$ (${\rm mF_1}$) and the mean Intersection over Union (${\rm mIoU}$) in our experiments. These widely adopted metrics enable a fair comparison of our method's performance against state-of-the-art methods. The ${\rm mF_1}$ and ${\rm mIoU}$ are calculated for the five foreground classes, while the class labeled as {\em Clutter} is treated as a background or sparse class, and hence, performance statistics for this class are not reported \cite{wu2023cmtfnet, ma2024sam}. The ${\rm F_1}$ and ${\rm IoU}$ for each class indexed by $n$ are computed using the following formulas:
\begin{eqnarray}
{\rm F_1}&=&2\times\frac{p_{n}r_{n}}{p_{n}+r_{n}},\\\label{eq_metric1}
{\rm IoU}&=&\frac{TP_{n}}{TP_{n}+FP_{n}+FN_{n}},\label{eq_metric2}
\end{eqnarray}
where $TP_{n}$, $FP_{n}$, and $FN_{n}$ are true positives, false positives, and false negatives for the $n$-th class, respectively. Furthermore, $p_{n}$ and $r_{n}$ are given by:
\begin{eqnarray}
p_{n}&=&\frac{TP_{n}}{TP_{n}+FP_{n}},\\\label{eq_metric3}
r_{n}&=&\frac{TP_{n}}{TP_{n}+FN_{n}}.\label{eq_metric4}
\end{eqnarray}
After calculating the ${\rm F_1}$ and ${\rm IoU}$ for the foreground classes based on the definitions provided, their mean values, referred to as ${\rm mF_1}$ and ${\rm mIoU}$, can be obtained, respectively.

\subsection{Implementation details}
The experiments were conducted using PyTorch on a single NVIDIA GeForce RTX 4090 GPU with 24GB of RAM. All models were trained with the stochastic gradient descent algorithm, utilizing a learning rate of $0.01$, a momentum of $0.9$, and a weight decay coefficient of $0.0005$. The total number of training epochs was set to $50$, and the multi-step learning rate schedule was applied with step values of $[25, 35, 45]$ and a gamma of $0.1$. The batch size was set to $10$. Simple data augmentations, such as random rotation and flipping, were applied after the sliding window collected the samples. The deep feature refinement module consists of $4$ stacked DeepKAN modules, i.e., ${\rm N} = 4$. To demonstrate the flexibility and robustness of the proposed DeepKANSeg, we performed experiments using two distinct pre-trained backbones: CNN-based ResNet18 and Vision Transformer-based ViT-L. For the latter, we utilized SAM's encoder \cite{kirillov2023segment} and fine-tuned it using Adapter \cite{wu2023medical}, which can greatly reduce training costs. The differences in structure and model scale allowed us to comprehensively evaluate the effectiveness of the method. Finally, all models in this study are optimized using the cross-entropy loss, without the introduction of any additional loss functions, to ensure the fairness of the comparative experiments.

\begin{table*}[t]\footnotesize
	\centering
	\caption{Quantitative comparison results on the ISPRS Vaihingen dataset with state-of-the-art methods. Here we present the accuracy of each category in ${\rm F_1}$ and ${\rm IoU}$ format. \textbf{Bold} values denote the best performance.}
	  \renewcommand\arraystretch{1.2}
		\begin{tabular}{ccccccccc}
			\hline
			\textbf{Method} & \textbf{Backbone} &  \textbf{impervious surface} & \textbf{building} & \textbf{low vegetation}  & \textbf{tree}  & \textbf{car}  & \textbf{mF1}    & \textbf{mIoU} \\
			\hline
			ABCNet  \cite{li2021abcnet}  &ResNet-18& 89.78/81.45 & 94.30/89.21 & 78.49/64.59 & 90.08/81.95 & 74.05/58.80 & 85.34 & 75.20 \\
			TransUNet \cite{chen2021transunet}   &R50-ViT-B& 90.77/83.10 & 94.32/89.25 & 79.02/65.32 & 90.53/82.70 & 82.66/70.45 & 87.46 & 78.16 \\
			UNetformer \cite{wang2022unetformer} &ResNet-18& 92.33/85.76 & 96.25/92.78 & 80.47/67.33 & 90.85/83.22 & 89.35/80.75 & 89.85 & 81.97 \\
			FTUNetformer \cite{wang2022unetformer} &Swin-Base& 93.41/87.64 & 96.92/94.02 & 81.53/68.82 & 90.91/83.33 & 88.46/79.31 & 90.24 & 82.62 \\
			CMTFNet    \cite{wu2023cmtfnet}   &ResNet-50& 92.53/86.09 & 96.95/94.09 & 79.98/66.64 & 90.22/82.19 & 89.87/81.60 & 89.91 & 82.12 \\
			SSNet  \cite{yao2024ssnet}  &ViT-B& 92.06/86.08 & 96.79/93.79 & 80.53/67.40 & 91.07/83.60 & 88.10/78.73 & 89.87 & 82.04 \\
			RS$^{3}$Mamba \cite{ma2024rs}    &R18-MambaT& 92.83/86.62 & 96.82/93.83 & 80.84/67.84 & 91.10/83.66 & 90.09/81.97 & 90.34 & 82.78 \\
			\hline
			\multirow{2}{*}{DeepKANSeg} &ResNet-18& 92.90/86.74 & 96.89/93.97 & 80.58/67.48 & 90.99/83.47 & 90.68/82.94 & 90.41 & 82.92 \\
			 &ViT-L& \textbf{93.46/87.72} & \textbf{97.03/94.24} & \textbf{82.93/70.83} & \textbf{92.09/85.34} & \textbf{90.18/82.11} & \textbf{91.14} & \textbf{84.05} \\
			\hline
	\end{tabular}\label{tab:vlist}
\end{table*}

\begin{figure*}[h!]
\centering
{\includegraphics[width=0.95\linewidth]{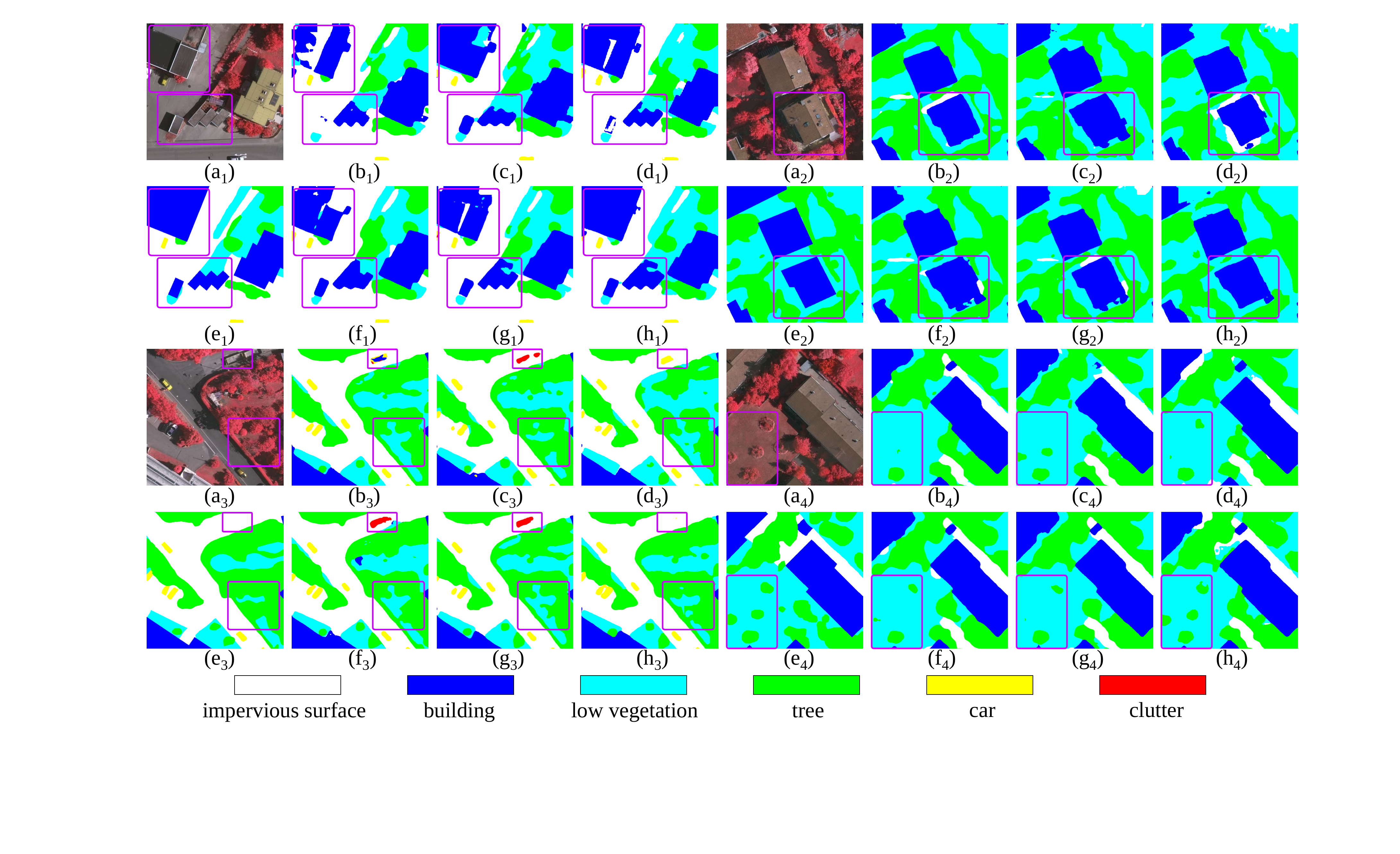}}
\caption{Visualization results on the ISPRS Vaihingen dataset with the size of $512 \times 512$. (a) NIRRG images, (b) UNetformer, (c) FTUNetformer, (d) CMTFNet, (e) Ground Truth, (f) SSNet, (g) RS$^{3}$Mamba, (h) The proposed DeepKANSeg. Some purple boxes are marked to highlight the differences.}
\label{fig_v_compare}
\end{figure*}

\subsection{Performance Comparison}
\begin{figure*}[h!]
\centering
{\includegraphics[width=0.95\linewidth]{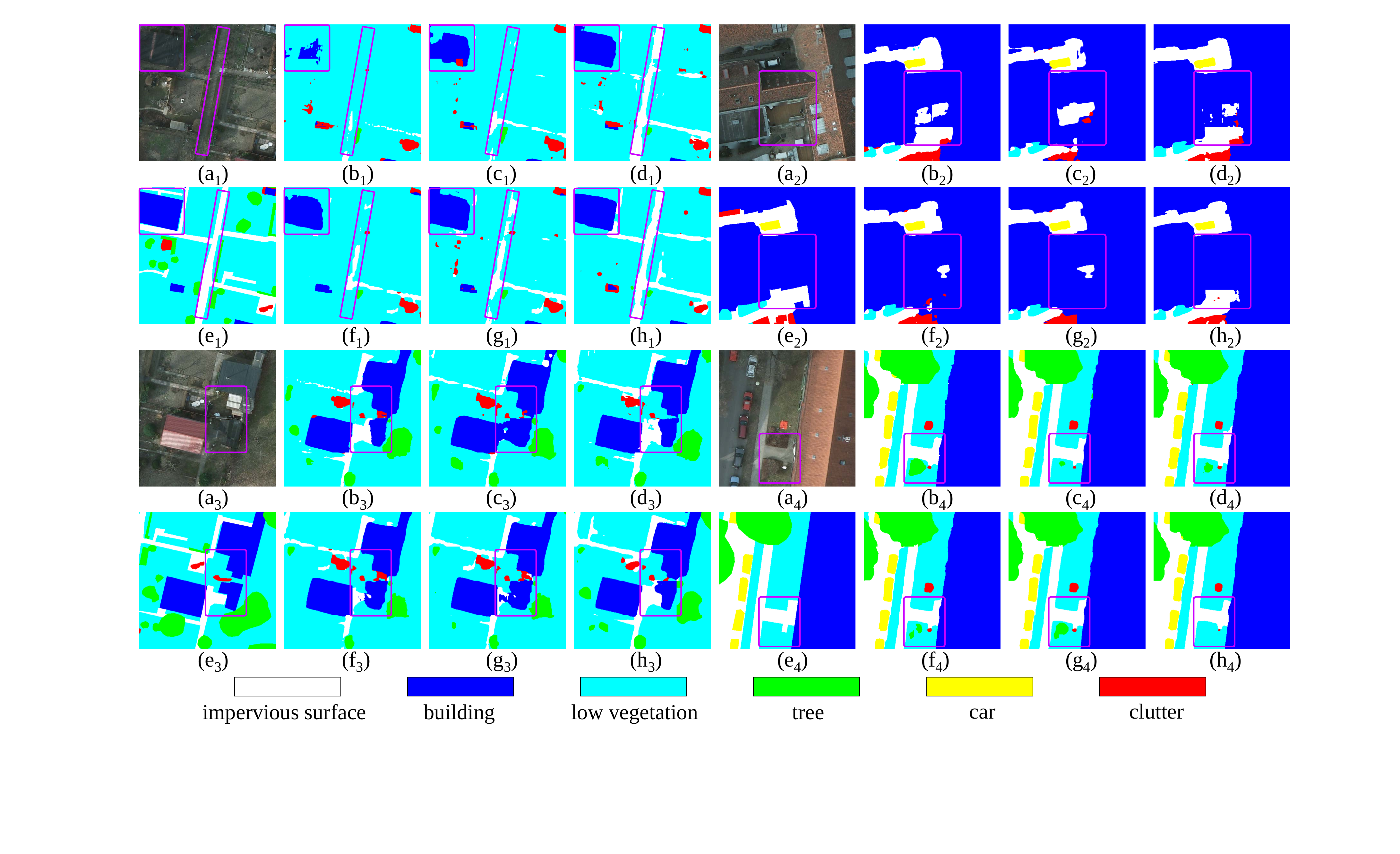}}
\caption{Visualization results on the ISPRS Potsdam dataset with the size of $512 \times 512$. (a) RGB images, (b) TransUNet, (c) UNetformer, (d) CMTFNet, (e) Ground Truth, (f) SSNet, (g) RS$^{3}$Mamba, (h) The proposed DeepKANSeg. Some purple boxes are marked to highlight the differences.}
\label{fig_p_compare}
\end{figure*}

We benchmarked the performance of the proposed DeepKANSeg against seven representative state-of-the-art methods, namely ABCNet \cite{li2021abcnet}, TransUNet \cite{chen2021transunet}, UNetFormer \cite{wang2022unetformer}, FTUNetFormer \cite{wang2022unetformer}, CMTFNet \cite{wu2023cmtfnet}, SSNet \cite{yao2024ssnet}, and RS$^{3}$Mamba \cite{ma2024rs}. These methods cover a range of widely used architectures, including CNN, Transformer, and Mamba \cite{zhu2024vision}, allowing for a comprehensive validation of our approach across various classical network architectures. The quantitative results for DeepKANSeg and the comparative methods are summarized in Table~\ref{tab:vlist} and Table~\ref{tab:plist}. Some visualization results are presented in Fig. \ref{fig_v_compare} and Fig. \ref{fig_p_compare}.

\begin{table*}[t]\footnotesize
	\centering
	\caption{	Quantitative comparison results on the ISPRS Potsdam dataset with state-of-the-art methods. Here we present the accuracy of each category in ${\rm F_1}$ and ${\rm IoU}$ format. \textbf{Bold} values denote the best performance.}
	  \renewcommand\arraystretch{1.2}
		\begin{tabular}{ccccccccc}
			\hline
			\textbf{Method} & \textbf{Backbone} &  \textbf{impervious surface} & \textbf{building} & \textbf{low vegetation}  & \textbf{tree}  & \textbf{car}  & \textbf{mF1}    & \textbf{mIoU} \\
			\hline
			ABCNet  \cite{li2021abcnet}  &ResNet-18& 90.05/81.90 & 94.53/89.62 & 81.87/69.31 & 79.61/66.12 & 92.30/85.70 & 87.67 & 78.51 \\
			TransUNet \cite{chen2021transunet}   &R50-ViT-B& 92.06/85.28 & 95.95/92.22 & 85.05/73.99 & 85.28/74.33 & 95.13/90.71 & 90.69 & 83.31 \\
			UNetformer \cite{wang2022unetformer} &ResNet-18& 92.40/85.87 & 96.54/93.32 & 85.34/74.43 & 86.07/75.55 & 95.59/91.56 & 91.19 & 84.15 \\
			FTUNetformer \cite{wang2022unetformer} &Swin-Base& 88.64/79.60 & 93.25/87.36 & 80.74/67.70 & 76.81/62.35 & 92.29/85.68 & 86.35 & 76.54 \\
			CMTFNet    \cite{wu2023cmtfnet}   &ResNet-50& 92.46/85.97 & 96.81/93.82 & 86.30/75.90 & \textbf{86.95/76.91} & 96.11/92.51 & 91.73 & 85.02 \\
			SSNet  \cite{yao2024ssnet}  &ViT-B& 91.98/85.16 & 96.58/93.38 & 85.32/74.40 & 85.93/75.33 & 95.78/91.91 & 91.12 & 84.02\\
			RS$^{3}$Mamba \cite{ma2024rs}    &R18-MambaT& 92.37/85.83 & 96.87/93.94 & 85.42/74.55 & 86.23/75.79 & 95.90/92.12 & 91.36 & 84.45 \\
			\hline
			\multirow{2}{*}{DeepKANSeg} &ResNet-18& 92.40/85.87 & 97.06/94.29 & 86.24/75.80 & 85.75/75.05 & 95.90/92.13 & 91.47 & 84.63 \\
			 &ViT-L& \textbf{92.69/86.37} & \textbf{97.11/94.39} & \textbf{86.84/76.74} & 86.89/76.83 & \textbf{96.30/92.87} & \textbf{91.97} & \textbf{85.44} \\
			\hline
	\end{tabular}\label{tab:plist}
\end{table*}

\subsubsection{Performance Comparison on the Vaihingen dataset}
As presented in Table~\ref{tab:vlist}, the proposed DeepKANSeg demonstrates significant advancements in both ${\rm F_1}$ and ${\rm IoU}$ metrics compared to comparative methods. Specifically, DeepKANSeg (ResNet-18) achieved notable improvements of $0.56\%$ in ${\rm mF_1}$ and $0.95\%$ in ${\rm mIoU}$ over the baseline UNetformer. Notably, DeepKANSeg delivered substantial gains in individual classes, with improvements of $0.98\%$ on the {\em Imp.} class, $1.19\%$ on the {\em Bui.} class, and $2.19\%$ on the {\em Car} class in terms of ${\rm IoU}$. These enhancements indicate that the proposed model effectively exploits and leverages detailed information for more accurate category predictions. When compared to state-of-the-art models, DeepKANSeg (ViT-L) outperformed across all categories. For instance, the ${\rm F_1}$ and ${\rm IoU}$ for the {\em Low.} class improved by $1.4\%$ and $2.01\%$, respectively, compared to FTUNetformer, while the ${\rm F_1}$ and ${\rm IoU}$ for the {\em Tre.} class improved by $0.99\%$ and $1.68\%$, respectively, compared to RS$^{3}$Mamba. These improvements highlight the unique strengths of DeepKANSeg in various scenarios. First, with a unified backbone, the KAN-based module effectively captures finer details of ground objects, particularly excelling in segmenting well-defined and regular structures such as the {\em Imp.}, {\em Bui.}, and {\em Car} classes. Second, with a more advanced backbone like ViT-L, DeepKANSeg demonstrates superior performance across all metrics, with particularly significant gains in challenging categories such as {\em Low.} and {\em Tre.}. While simpler categories are relatively easier to segment, allowing other methods to perform well on certain individual classes, DeepKANSeg excels in extracting nuanced details, providing a robust and reliable foundation for accurately segmenting complex categories. Overall, the proposed DeepKANSeg (ViT-L) achieved an impressive ${\rm mF_1}$ of $91.14\%$ and ${\rm mIoU}$ of $84.05\%$, representing increases of $0.8\%$ and $1.27\%$ compared to the corresponding metrics of the best comparative method, RS$^{3}$Mamba. These improvements can be attributed to the innovative integration of the KAN-based deep feature refinement module and the GLKAN decoder, which enable more effective extraction and refinement of high-dimension and multi-scale semantic features. In sharp contrast to the traditional methods that may struggle with complex spatial patterns in remote sensing scenes, DeepKANSeg excels in capturing detailed context, thereby addressing the challenges posed by high inter-class similarity and fine-grained intra-class variations. Furthermore, the consistent enhancement across all categories highlights the robustness and generalization ability of the proposed method. These results not only validate the effectiveness of DeepKANSeg but also demonstrate its potential for advancing semantic segmentation in complex and diverse ground environments.

Fig.~\ref{fig_v_compare} presents a visual comparison of the results obtained by the five best methods under consideration. Remote sensing images exhibit greater complexity than natural images, with buildings varying significantly in scale, characterized by neat borders but diverse shapes. Trees and low vegetation are often intertwined, further increasing the complexity. The proposed DeepKANSeg demonstrates superior performance by capturing intricate edges with smoother results, producing more complete and connected objects. Specifically, the deep feature refinement module excels in accurately recognizing complex, long-range semantic relationships, aiding in the identification of complete objects. Furthermore, the GLKAN modules effectively extract integrated global-local information, providing a refined foundation for final predictions. These advantages enable DeepKANSeg to deliver more precise classifications compared to other methods. Highlighted in the purple boxes across all subfigures of Fig.~\ref{fig_v_compare}, DeepKANSeg successfully identifies buildings, trees, and low vegetation surrounded by other ground objects. These results in cleaner and more complete segmentations, further emphasizing its effectiveness in handling the complexity of remote sensing images.

\subsubsection{Performance Comparison on the Potsdam dataset}
Experiments on the ISPRS Potsdam dataset yielded results consistent with those observed on the ISPRS Vaihingen dataset, despite differences in geographical location, spectral channels, and sampling resolutions. As shown in Table~\ref{tab:plist}, DeepKANSeg (ResNet-18) achieved ${\rm mF_1}$ and ${\rm mIoU}$ of $91.47\%$ and $84.63\%$, respectively, representing improvements of $0.28\%$ and $0.48\%$ over the baseline UNetformer. Significant gains were observed in three classes, with ${\rm IoU}$ improvements of $0.97\%$ for the {\em Bui.} class, $1.37\%$ for the {\em Low.} class, and $0.57\%$ for the {\em Car} class. For the proposed DeepKANSeg (ViT-L), the ${\rm mF_1}$ and ${\rm mIoU}$ were $91.97\%$ and $85.02\%$, which corresponds to increases of $0.24\%$ and $0.42\%$, respectively, over CMTFNet. Notably, substantial improvements were observed in the segmentation of {\em Imp.}, {\em Bui.}, {\em Low.}, and {\em Car} classes compared to other state-of-the-art methods. Fig.~\ref{fig_p_compare} illustrates visualization examples from the ISPRS Potsdam dataset, with areas of interest marked by purple boxes in all subfigures. Similar to the results on Vaihingen, DeepKANSeg's predictions exhibit reduced noise points and more reliable segmentation, especially in challenging scenes.

\begin{table}[h]
	\centering
	\caption{Ablation study of the proposed approach on two different backbones.}
		\begin{tabular}{c|cc|cc}
			\hline
			\textbf{Backbone} & \textbf{DeepKAN}   &  \textbf{GLKAN}      & \textbf{mF1(\%)}  & \textbf{mIoU(\%)}   \\
			\hline
			\multirow{4}{*}{ResNet-18}  &   &      &  89.85   & 81.97  \\
			 &\checkmark   &      &  90.27   & 82.68  \\
		  &&  \checkmark    &    90.18   & 82.51 \\
			&\checkmark   &  \checkmark    & \textbf{90.41}    & \textbf{82.92} \\
			\hline
			\multirow{4}{*}{ViT-L} &   &      &  90.61   & 83.21  \\
			&\checkmark   &      &  90.95   & 83.74  \\
		  &&  \checkmark    &    90.79   & 83.44 \\
			&\checkmark   &  \checkmark    & \textbf{91.14}    & \textbf{84.05} \\
			\hline
	\end{tabular}\label{tab:ablalist}
\end{table}

\begin{figure}[h]
\centering
{\includegraphics[width=1\linewidth]{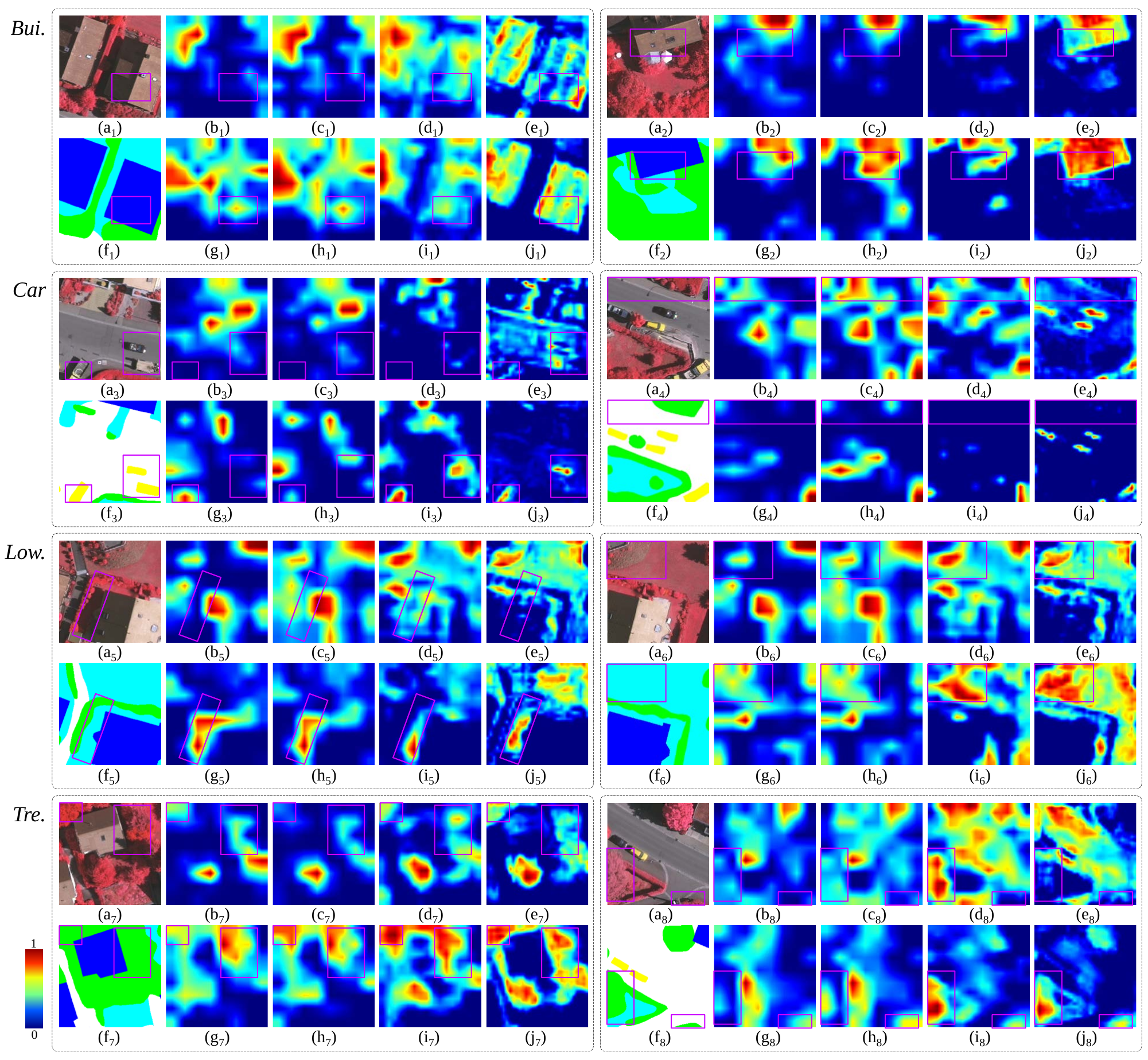}}
\caption{Eight groups of heatmap samples are presented. In each subfigure: (a$_{\star}$) represents the NIIRG image, (b$_{\star}$–e$_{\star}$) depict four heatmaps generated by UNetformer, (f$_{\star}$) shows the Ground Truth, and (g$_{\star}$–j$_{\star}$) illustrate four heatmaps produced by DeepKANSeg (ResNet-18), where $\star \in \{1,2,3,4,5,6,7,8\}$ corresponds to eight distinct samples. The first, second, third, and fourth pairs of samples demonstrate how the two models determine whether a pixel belongs to {\em Bui.}, {\em Car}, {\em Low.}, and {\em Tre.}, respectively. Purple boxes have been added to emphasize the key differences.}
\label{fig_heatmap}
\end{figure}

\subsection{Ablation Study}
To assess the effectiveness of each proposed component in DeepKANSeg, ablation experiments were performed with two different backbones while preserving the encoder-decoder framework. As outlined in Table~\ref{tab:ablalist}, these experiments were designed to examine the specific contributions of the KAN-based modules. Initially, the baselines were established in the first rows. When using ResNet-18 as the backbone, the baseline corresponds to UNetFormer. For the second configuration, the encoder was replaced with ViT-L, and an Adapter-based fine-tuning strategy was applied, enabling the encoder to efficiently extract features from remote sensing images. In the second row of each configuration, DeepKAN modules were introduced after the encoder to evaluate their individual effect, while in the third row of each configuration, the decoder was upgraded to the GLKAN-based version to evaluate its contribution. These two configurations provide insights into the distinct roles played by the DeepKAN and GLKAN modules. The final rows represent the complete DeepKANSeg networks, integrating both DeepKAN and GLKAN modules.

Inspection of Table~\ref{tab:ablalist} reveals that both the DeepKAN and GLKAN modules are essential for the improved performance of the proposed DeepKANSeg, with the contribution of DeepKAN being particularly pronounced. This highlights the critical role of deep feature refinement in remote sensing tasks. Furthermore, by effectively extracting and refining high-dimensional features, the subsequent semantic recovery process becomes more streamlined and accurate. The ablation experiments also validate the rationale behind the framework designation. DeepKAN leverages its KAN-based fine-grained learning and stacked structure to refine abstract high-dimensional semantic features, proving highly effective in capturing the intricate details and complexities characteristic of remote sensing images. Meanwhile, the GLKAN-based decoder enables comprehensive exploitation of global-local features, ensuring more precise predictions for objects with diverse spatial scales. The complementary roles of these modules are evident, and their combination ensures the robustness of DeepKANSeg, achieving superior performance in remote sensing image semantic segmentation tasks.

To clearly illustrate the effectiveness of the DeepKAN and GLKAN modules, Fig.~\ref{fig_heatmap} presents heatmaps generated by the baseline UNetformer and the proposed DeepKANSeg (ResNet-18). For UNetformer, heatmaps were captured after the encoder and each of the three decoder blocks, while for DeepKANSeg, corresponding heatmaps were obtained from equivalent layers. The four rows of heatmaps demonstrate how the two models identify a pixel belonging to the classes {\em Bui.}, {\em Car}, {\em Low.}, and {\em Tre.}, respectively. First, it is evident that our method produces larger and more precise regions with high scores corresponding to the target classes, as observed across most samples. This demonstrates that DeepKANSeg effectively extracts category-specific features more comprehensively and thoroughly through refined learning. In addition, our model exhibits a stronger capability to discern differences between classes. For example, in the highlighted box of the fourth sample, our approach avoids focusing on objects that are irrelevant to the target class. This indicates that the model possesses a robust resistance to noise and distractions. These capabilities stem from the refinement power of the KAN-based modules, which substantially enhances performance in remote sensing image semantic segmentation tasks.

\begin{table*}[t]\small
	\centering
	\caption{Model complexity analysis performed with two $256 \times 256$ images on a single NVIDIA GeForce RTX 4090 GPU. The ${\rm mIoU}$ values correspond to the results obtained on the ISPRS Vaihingen dataset. \textbf{Bold} values denote the best performance.}
	\setlength{\tabcolsep}{5mm}{
		\begin{tabular}{m{2.6cm}<{\centering}|m{1.8cm}<{\centering}|m{0.8cm}<{\centering}m{0.8cm}<{\centering}m{0.8cm}<{\centering}m{0.8cm}<{\centering}m{0.8cm}<{\centering}}
			\hline
			\textbf{Model}   & \textbf{Backbone}  & \textbf{FLOPs (G)}  &  \textbf{Parameter (M)} & \textbf{Memory (MB)}  &  \textbf{Speed (FPS)}  &  \textbf{MIoU (\%)}  \\
			\hline
			ABCNet \cite{li2021abcnet} &ResNet-18& 7.81 & 13.67 & \textbf{838} & \textbf{477.76} & 75.20 \\
			TransUNet \cite{chen2021transunet} &R50-ViT-B& 64.55 & 105.32 & 1784 & 118.06 & 78.16 \\
			UNetformer \cite{wang2022unetformer} &ResNet-18& \textbf{5.87} & \textbf{11.69} & 842 & 382.97 & 81.97 \\
			FTUNetformer \cite{wang2022unetformer}  &Swin-Base& 50.84 & 96.14 & 2540 & 85.91 & 82.62 \\
			CMTFNet \cite{wu2023cmtfnet}   &ResNet-50& 17.14 & 30.07 & 1496 & 207.43 & 82.12 \\
			SSNet \cite{yao2024ssnet}   &ViT-B& 45.22 & 85.13 & 1824 & 26.48 & 82.04 \\
			RS$^{3}$Mamba \cite{ma2024rs}   &R18-MambaT& 28.25 & 43.32 & 1824 & 106.54 & 82.78 \\
			\hline
			\multirow{2}{*}{DeepKANSeg} &ResNet-18& 19.08 & 43.50 & 1476 & 122.59 & 82.92 \\
			&ViT-L& 16.86 & 56.67 & 3790 & 29.17 & \textbf{84.05} \\
			\hline
	\end{tabular}}\label{tab:scale}
\end{table*}
		
\subsection{Model Complexity Analysis}\label{sec:complexity}
The model complexity of the proposed method and comparative methods are evaluated using several key metrics: floating point operations (FLOPs), model parameters, memory footprint, and running speed. FLOPs measure the computational complexity of the model, while model parameters and memory footprint evaluate the model's scale and memory requirements, respectively. Running speed quantifies the time cost during both the training and inference stages. Ideally, an efficient method achieves lower FLOPs, fewer model parameters, reduced memory footprint, and faster running speed. It is worth noting that when DeepKANSeg employs ViT-L as the backbone, only the fine-tuned components are included in the complexity and parameter calculations, excluding the fixed parts.

Table~\ref{tab:scale} presents the complexity analysis results for all methods considered in this work. Inspection of Table~\ref{tab:scale} reveals that while our proposed model introduces some additional complexity compared to the baseline UNetFormer, it achieves significant performance improvements. Specifically, DeepKANSeg (ResNet-18) exhibits FLOPs and memory usage comparable to CMTFNet, and parameters and running speed comparable to RS$^{3}$Mamba. This indicates that incorporating KAN-based modules has an impact similar to adding more stacked modules in ResNet or introducing a mamba-based auxiliary branch as seen in RS$^{3}$Mamba. However, in terms of ${\rm mIoU}$ performance, our approach outperforms both methods. When employing a larger backbone, such as ViT-L, our model loses its advantage in terms of complexity but demonstrates a substantial improvement in segmentation performance. This highlights that the KAN-based modules effectively leverage the richer features extracted by larger backbones. By offering DeepKANSeg with two different backbone configurations, we ensure that the proposed method provides superior scalability and segmentation performance, accommodating diverse application requirements compared to existing methods.

\subsection{Discussion}\label{sec:discussion}
This work introduces two KAN-based modules, namely DeepKAN and GLKAN, designed specifically for semantic segmentation in remote sensing images. As an initial exploration in this domain, we thoroughly evaluate their performance on two high-resolution datasets, conducting extensive analytical experiments to assess their effectiveness. The proposed DeepKANSeg offers a straightforward approach to integrating KAN technology into remote sensing tasks, paving the way for further advancements in the field. Meanwhile, several areas warrant further investigation to maximize the potential of this technology:
\begin{itemize}[leftmargin=*]
\item{\bf KAN-based Encoder}: Feature extraction is a critical component of remote sensing tasks. Existing methods often rely on pre-trained models, and there are currently no backbones based on KAN, making it impossible to construct an efficient pure KAN-based network. Future work should also focus on designing efficient KAN-based backbones to explore the advantages and disadvantages of pure KAN-based networks.
\item{\bf Complexity}: Current KAN implementations utilize B-Splines as the foundational units, which pose challenges for parallel processing on GPU. Exploring more efficient functional approximation units may significantly lower the computational complexity and enhance the operational efficiency of KAN-based modules, making them more suitable for resource-constrained environments.
\item{\bf Interpretability}: Interpretability remains a standout feature of KAN technology. However, when dealing with high-dimensional features, it becomes challenging to quantify the information across different channels effectively. Developing interpretability methods beyond visual tools like heatmaps could provide a deeper understanding of KAN-based deep models and make their decision-making processes more transparent.
\end{itemize}
We hope this work to inspire further advancements and exploration of KAN-based methods in remote sensing tasks.

\section{Conclusion}\label{sec:con}
This work presents a novel network named DeepKANSeg to remote sensing image semantic segmentation by introducing two KAN-based modules, namely DeepKAN and GLKAN. In particular, DeepKAN is proposed to exploit the deep features with rich geographical information, while GLKAN is introduced to recover the spatial relationship by exploiting the global-local information. The proposed KAN-based modules excel in refined learning, enabling precise and robust segmentation of complex and intricate remote sensing scenes. Comprehensive experiments on two high-resolution remote sensing datasets validate the effectiveness of our method, demonstrating notable improvements in segmentation accuracy compared to state-of-the-art methods. Moreover, the adaptability of DeepKANSeg across different backbones highlights its scalability and practicality for diverse applications.

Despite its promising results, it remains challenging in computational complexity and the reliance on pre-trained encoders, as well as the need for further exploration into interpretability for high-dimensional features. These limitations present opportunities for future research to optimize KAN-based designs and enhance their efficiency and transparency. Finally, this work provides a solid foundation for incorporating KAN-based technology into remote sensing tasks, opening new avenues for advancements in both methodological innovation and practical applications in the field.

\small
\bibliographystyle{IEEEtran}
\bibliography{references}

\end{document}